%% file: root.tex
\documentclass[letterpaper, 10 pt, conference]{ieeeconf}  % Comment this line out if you need a4paper

\IEEEoverridecommandlockouts                              % This command is only needed if 
\usepackage{graphics} % for pdf, bitmapped graphics files
\usepackage{epsfig} % for postscript graphics files
\usepackage{mathptmx} % assumes new font selection scheme installed
\usepackage{times} % assumes new font selection scheme installed
\usepackage{amsmath} % assumes amsmath package installed
\usepackage{amssymb}  % assumes amsmath package installed

\usepackage{multirow} 
\usepackage{booktabs} 
\usepackage{makecell}
\usepackage[T1]{fontenc}
\usepackage[dvipsnames]{xcolor}
\definecolor{LightRed}{rgb}{1.0, 0.6, 0.6}
\definecolor{LightBlue}{rgb}{0.68, 0.85, 0.9}
\usepackage{svg}

\title{\LARGE \bf
Learning Robotic Policy with Imagined Transition: Mitigating the Trade-off between Robustness and Optimality
}

\author{Wei Xiao, Shangke Lyu$^{*}$, Zhefei Gong, Renjie Wang, Donglin Wang$^{*}$% <-this % stops a space
\thanks{All the authors are with Machine Intelligence Lab (MiLAB), School of Engineering, Westlake University, Hangzhou 310024, China.}
\thanks{$^{*}$Corresponding author.}
}

\begin{document}

\maketitle
\thispagestyle{empty}
\pagestyle{empty}

%%%%%%%%%%%%%%%%%%%%%%%%%%%%%%%%%%%%%%%%%%%%%%%%%%%%%%%%%%%%%%%%%%%%%%%%%%%%%%%%
\begin{abstract}
\input{section/0_abs}

\end{abstract}

%%%%%%%%%%%%%%%%%%%%%%%%%%%%%%%%%%%%%%%%%%%%%%%%%%%%%%%%%%%%%%%%%%%%%%%%%%%%%%%%
\section{INTRODUCTION}
\input{section/1_intro}

\section{Related Work}
\input{section/2_related_work}

\section{Preliminary}
\input{section/3_preliminary}

\section{Method}
\input{section/4_method}

\section{Experiments}
\input{section/5_experiment}

\section{Conclusion}
\input{section/6_conclusion}

\section*{ACKNOWLEDGMENT}
This work was supported by the National Science and Technology Innovation 2030-Major Projects (Grant No. 2022ZD0208800) and the National Natural Science Foundation of China (Grant No. 62176215 and 62003018).

%%%%%%%%%%%%%%%%%%%%%%%%%%%%%%%%%%%%%%%%%%%%%%%%%%%%%%%%%%%%%%%%%%%%%%%%%%%%%%%%

\bibliographystyle{./IEEEtran}
\bibliography{references}

\addtolength{\textheight}{-12cm} 

\end{document}

%% file: section/0_abs.tex
Existing quadrupedal locomotion learning paradigms usually rely on extensive domain randomization to alleviate the sim2real gap and enhance robustness. It trains policies by introducing various disturbances into simulated environments to improve their adaptability under uncertainty. However, since optimal performance without disturbances often conflicts with the need to handle worst-case scenarios, this paradigm has a trade-off between optimality and robustness. This trade-off forces the learned policy to prioritize stability in diverse and challenging environments over efficiency and accuracy in no-disturbed ones, leading to overly conservative behaviors that sacrifice peak performance. Inspired by disturbance rejection control, we propose a two-stage framework that integrates policy learning with motion reference to mitigate this phenomenon. This framework enhances the conventional reinforcement learning (RL) approach by incorporating imagined transitions as reference inputs. The imagined transitions are derived from an optimal policy and a dynamics model operating within an idealized setting without disturbances. By providing this reference, we found that the policy could generate actions that are better aligned with the desired motion, which led to accelerated training and reduced tracking errors under external disturbances.

%% file: section/1_intro.tex
% Background
Learning-based locomotion control methods for quadruped robots have achieved remarkable results in recent years~\cite{xie2021dynamics, tan2018sim, kumar2021rma, nahrendra2023dreamwaq, margolis2023walk, long2024him}.
To adapt to various wild environments, these methods collected large amounts of data in massive parallel simulators with extensive dynamics variations via domain randomization~\cite{tobin2017domain}.
This data-driven paradigm allows RL-based robots to adapt to realistic terrain and unseen disturbances.

% Challenges
While domain randomization is a widely used technique to improve the robustness of robot policies, previous studies~\cite{pmlr-v100-mehta20a, tiboni2023doraemon} have shown that it often sacrifices optimality, resulting in conservative behaviors.
For instance, when trained with large payload variations, the learned policy tends to maintain a lower base height rather than achieve the target height specified in the reward function.
This example reveals that the robustness achieved through domain randomization essentially comes at the cost of compromising certain task objectives under unknown changes.
Consequently, it becomes difficult to consistently meet high task standards, which limits the applicability of domain randomization in complex tasks, particularly those requiring high accuracy.

\begin{figure}[h]
    \centering
    \includegraphics[width=1.0\linewidth]{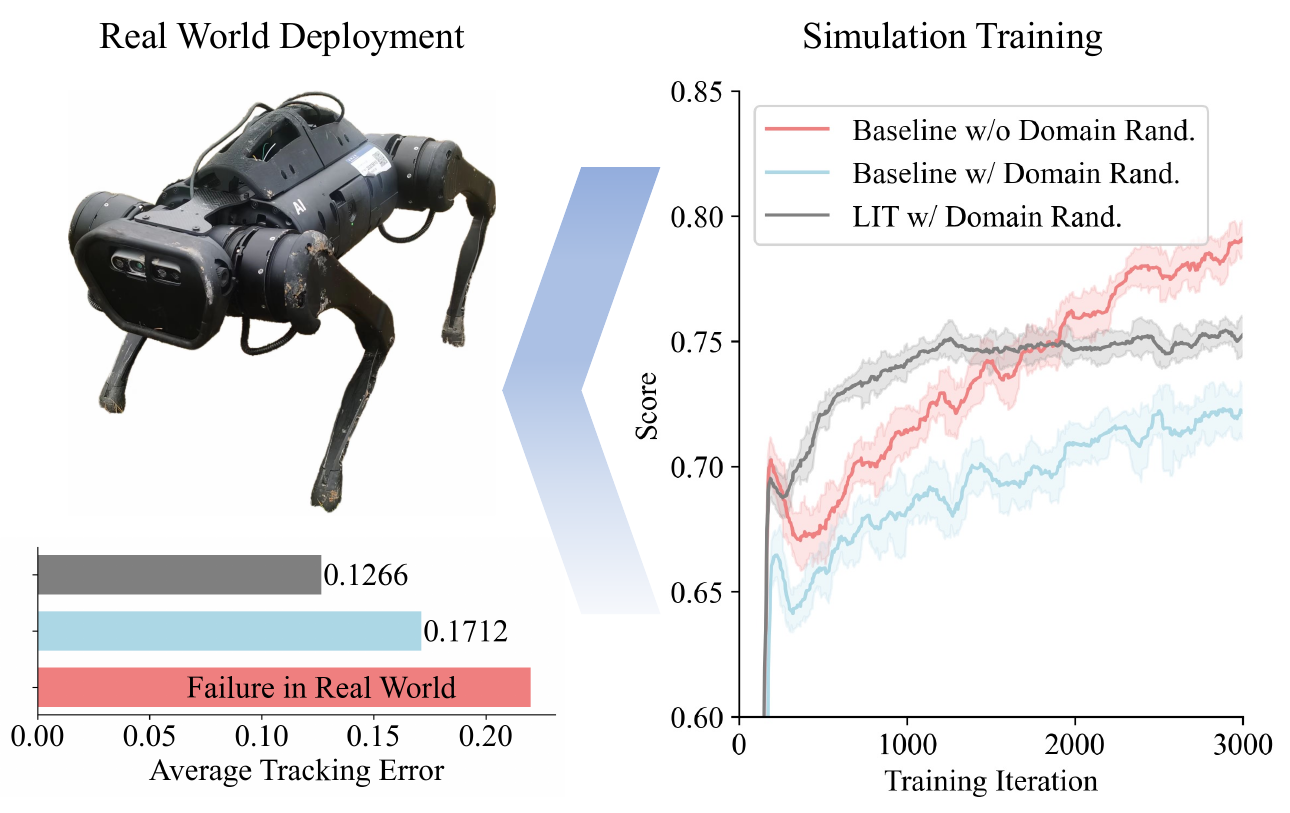}
    \caption{\textbf{Proposed LIT has lower tracking error than baseline.} For current RL-based locomotion paradigms (Baseline), domain randomization trades optimality for robustness.}
    \label{fig:intro}
    \vspace{-0.2in}
\end{figure}

% Why 
% This phenomenon is caused by the large amount of random disturbances introduced by domain randomization, which greatly increases the exploration space of RL policy training.
% Furthermore, the environment is modeled as a Partially Observable Markov Decision Process (POMDP)~\cite{spaan2012partially, kurniawati2022partially},  the agent lacks full access to all the perturbed environmental parameters.
% This partial observability forces the policy to adopt conservative behaviors to hedge against potential worst-case scenarios.

% Why Reference
To mitigate this problem, we require a method that can maintain task standards without compromising adaptability under disturbances.  
A key element for enabling robots to recover from unknown perturbations is the explicit identification of the goal that the policy should adapt to. 
This aligns with the control-theoretic perspective, particularly classical disturbance rejection control, where reference signals are used to compute counteractions that cancel perturbation effects
These reference signals define the desired system behavior and represent the performance the controller seeks to preserve despite disturbances.
In RL policy training, however, such references are only implicitly encoded in the reward functions, and no explicit goal-oriented guidance is provided.
Inspired by disturbance rejection control, we therefore propose to supply explicit motion references in RL policy learning.
These references not only provide a clearer objective for training, enabling the policy to learn to resist disturbances and track desired motions, but also deliver richer guidance for policy inference.

% Motivation
Normally, it is straightforward to train a policy in simulation under a fixed set of environmental parameters without any randomization.
Although such a policy cannot be directly deployed in reality due to limited robustness, it can still serve as a reference motion to guide both policy training and inference. Alternatively speaking, a well-trained policy in an ideal simulation can be leveraged to generate goal information that assists the deployed policy in making better decisions. 
To this end, we propose \textbf{LIT} (\textbf{L}earning with \textbf{I}magined \textbf{T}ransition), a two-stage framework training paradigm that enables the learning of robust locomotion policies from ideal motion. 
By introducing “ideal” actions and transitions, the policy could generate actions that are better aligned with nominal dynamics under disturbances (as illustrated by the gray line in Fig.~\ref{fig:intro}), thereby enabling more robust and accurate task execution.

% Framework Overview
Specifically, we reconstruct a learnable dynamics model to approximate transition propagation, obtained from a non-perturbative simulation environment together with an ideal policy. 
These provide reference actions and next-step ideal observations for policy training and inference, guiding the policy toward robust locomotion even under external disturbances.
To achieve this, we incorporate ideal transitions into the policy input while continuing to optimize the policy in an RL manner.
This approach not only improves training efficiency but also enhances the generalization capability of the locomotion policy. 
However, the predicted observation from the dynamics model and ideal policy may become ineffective in Out-of-Distribution (OOD) scenarios. 
If we include it in the optimization objective, it may lead to a training crash or non-convergence.
To address this, we designed an adjustment mechanism of the dynamics model to mitigate the negative impact of OOD observation during policy learning.

% Contribution
In summary, the contributions of our work are as follows:

$\bullet$ We propose \textbf{LIT}, a two-stage RL-based locomotion paradigm that consists of motion reference learning and robust policy learning, leveraging ideal transitions obtained through joint training of the policy and dynamics model in a non-perturbative simulation environment.

$\bullet$ We design an adjustment mechanism to mitigate the adverse effects of dynamics model errors when encountering out-of-distribution (OOD) observations, ensuring efficient and stable policy training.

$\bullet$ We conduct extensive experiments and ablations in both simulators and real-world scenarios. The empirical results demonstrate that LIT improves training efficiency and significantly enhances quadruped locomotion performance under unknown disturbances.

%% file: section/2_related_work.tex
\subsection{RL-based Legged Locomotion}
Reinforcement learning has seen significant progress in quadrupedal locomotion control, largely driven by advances in physics simulators like Mujoco~\cite{todorov2012mujoco} and IsaacGym~\cite{makoviychuk2021isaac}. This high-fidelity, GPU-accelerated platform allows for efficient RL training by simulating complex environments~\cite{rudin2022learningwalkminutesusing}.  

To mitigate the uncertainty caused by partial observation, the teacher-student framework~\cite{kumar2021rma, doi:10.1126/scirobotics.abc5986} was proposed, enabling quadrupedal robots to adapt to unknown terrains and greatly improving locomotion flexibility and robustness. Much of the work~\cite{pmlr-v205-agarwal23a, Cheng2024parkour, kumar2022adapting} has continued in this structure with impressive results.
Walk-These-Way~\cite{margolis2023walk} proposes a method to improve the generalization of quadrupedal locomotion by training the robot in diverse environments.
Another attempt is introducing environmental representation, such as~\cite{ji2022concurrent, nahrendra2023dreamwaq, long2024him}.
DreamWaQ~\cite{nahrendra2023dreamwaq}, leverages variational autoencoders (VAE)~\cite{higgins2016beta} to learn environmental representations, while HIMloco~\cite{long2024him} employs contrastive learning techniques to learn terrain representations.
Recent advances~\cite{lyu2023composite, lyu2024rl2ac} introduce model-based control~\cite{kim2019highly} and adaptive control~\cite{ioannou1996robust} to improve the interpretability and robustness of RL-based locomotion.
To address the challenges of the sim2real gap, domain randomization has been commonly used~\cite{xie2021dynamics, tan2018sim, kumar2021rma, nahrendra2023dreamwaq, long2024him}.
It is a class of methods in which the policy is trained with a wide range of environment parameters and sensor noise to learn behaviors that are robust in this range. However, domain randomization trades optimality for robustness, sometimes leading to an over-conservative policy.

\subsection{Model-based RL}
Model-based Reinforcement Learning (MBRL) incorporates a learned dynamics model of the environment, which predicts future states given the current state and action. This allows MBRL methods to plan and use collected data efficiently. Popular frameworks include Dreamer series~\cite{hafner2019dream, hafner2020mastering, hafner2023mastering}, which uses a learned world model to perform planning in latent space, and TDMPC series~\cite{hansen2022temporal, hansen2023td}, which integrates model predictive control to improve sample efficiency.
In the legged locomotion tasks, recent MBRL approaches have demonstrated the power of learned dynamics models. DayDreamer~\cite{wu2023daydreamer}, inspired by Dreamer, enables effective planning and control for robotic locomotion. PIP-loco~\cite{shirwatkar2024pip} also adopts a model-based approach and combines it with model predictive control, resembling TDMPC. These methods utilize the learned model for trajectory planning or representation learning, leveraging the learned dynamics model to enhance performance.
In contrast to these works, our approach utilizes the dynamics model not for planning or representation, but as a reference to predict the next state, thereby more directly affecting policy learning.

%% file: section/3_preliminary.tex
\begin{figure*}[t]
    \centering
    \includegraphics[width=1\linewidth]{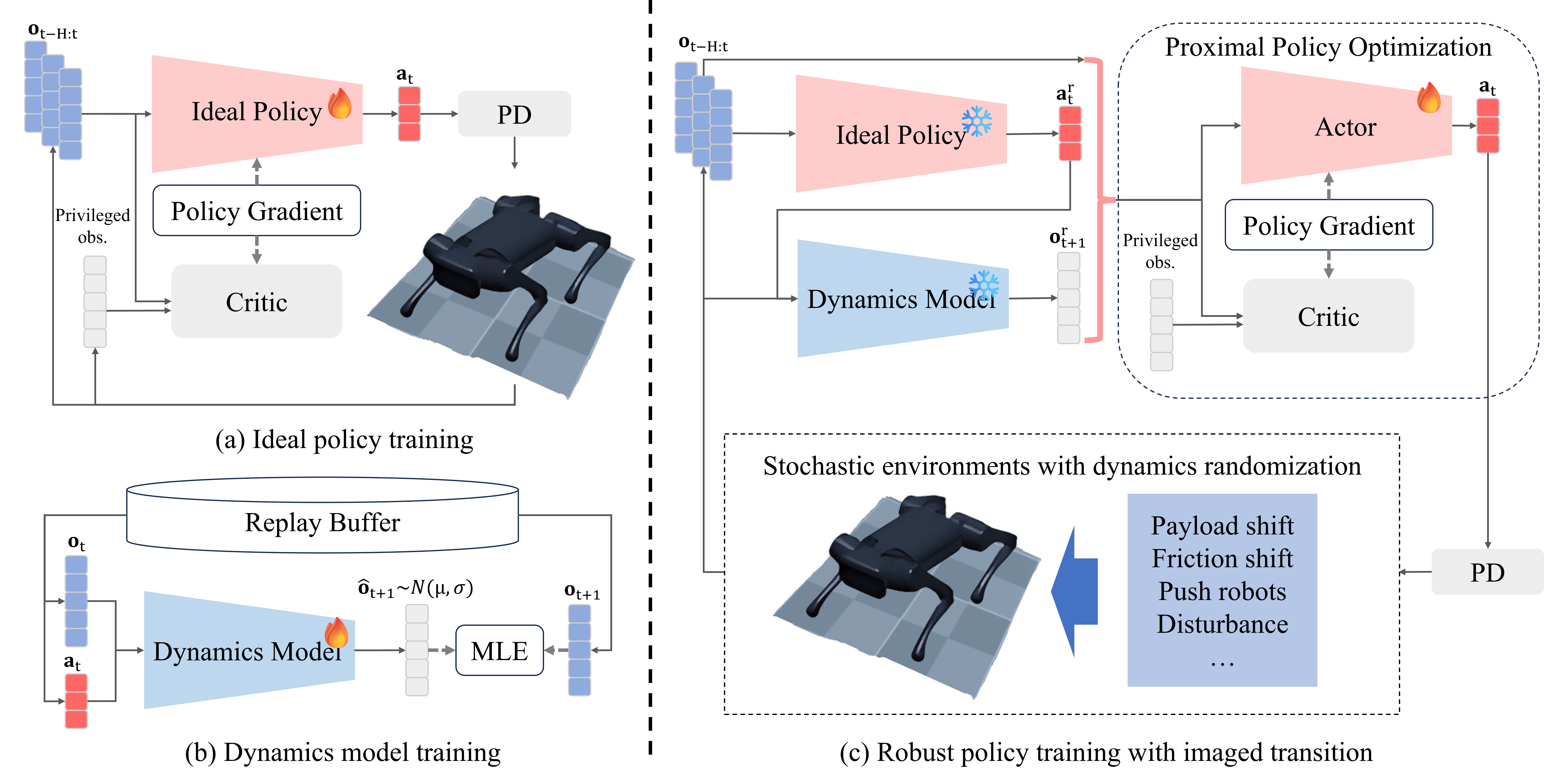}
    \caption{\textbf{Framework Overview of LIT.} The left part is motion reference learning in fixed dynamics. The right part is policy learning with imagined transition. $\mathbf{a}^{r}_{t}$ and $\mathbf{o}^{r}_{t+1}$ mean reference action and imagined next observation respectively. $[\mathbf{o}_{t}, \mathbf{a}^{r}_{t}, \mathbf{o}^{r}_{t+1}]$ constitutes the imagined transition}
    \label{fig:framework}
\end{figure*}

\paragraph{POMDP} 
In this study, the environment is modeled as an infinite-horizon partially observable Markov decision process (POMDP), represented by the tuple \( \mathcal{M} = (\mathcal{S}, \mathcal{O}, \mathcal{A}, d_0, p, r, \gamma) \). The full state \( s \in \mathcal{S} \), partial observation \( o \in \mathcal{O} \), and action \( a \in \mathcal{A} \) are all continuous variables. The system begins with an initial state distribution, denoted by \( d_0(s_0) \), and evolves according to the state transition probability \( p(s_{t+1} | s_t, a_t) \). Each state transition yields a reward determined by the reward function \( r: \mathcal{S} \times \mathcal{A} \to \mathcal{R} \), while the discount factor is specified by \( \gamma \in [0, 1) \).
Additionally, we define a temporal observation at time \( t \) over the previous \( H \) time steps as \( \mathbf{o}_{t-H:t} = \left[ \mathbf{o}_t \quad \mathbf{o}_{t-1} \quad \cdots \quad \mathbf{o}_{t-H} \right]^T \). This representation captures the history of partial observations within the temporal window, which is crucial for decision-making in partially observable settings.

\paragraph{State Space}  
The policy network takes partial observations \(\mathbf{o}_{t}\) as input, which consist of the desired velocity, proprioceptive data from the joint encoders and IMU, as well as the previous action \(\mathbf{a}_{t-1}\). The IMU provides the base angular velocity \(\mathbf{\omega}_t\) and the gravity direction in the robot's frame of reference, denoted by \(\mathbf{g}_t\). The desired velocity, \(\mathbf{c}_t = [v_x^c, v_y^c, \omega_{\text{yaw}}^c]\), represents the linear velocities in the longitudinal and lateral directions, as well as the angular velocity around the yaw axis. The joint encoders supply the joint position \(\mathbf{\theta}_t\) and the joint velocity \(\dot{\mathbf{\theta}}_t\).
In the training phase, the critic network is granted access to privileged information, enabling it to provide more accurate state value estimations. The input to the critic network, \(\mathbf{o}_{t}^{c}\), includes three additional components compared to the actor network's input \(\mathbf{o}_{t}^{a}\): the current velocity \(\mathbf{v}_t\), the external force \(\mathbf{f}_t\), and the surrounding ground height \(\mathbf{h}_t\).

% \begin{equation}
%   \mathbf{o}_t=\begin{bmatrix}\boldsymbol{\omega}_t & \mathbf{g}_t & \mathbf{c}_t & \boldsymbol{\theta}_t & \boldsymbol{\dot{\theta}}_t & \mathbf{a}_{t-1} \end{bmatrix}^T,
%     \label{eq:actor_obs}
% \end{equation}
% \begin{equation}
%     \mathbf{o}_{t}^{c} = \begin{bmatrix}{\mathbf{o}_{t}, \mathbf{v}_{t}, \mathbf{f}_{t}, \mathbf{h}_{t}}\end{bmatrix}^T,
%     \label{eq:critic_obs}
% \end{equation}

\paragraph{Action Space}  
The movement of each actuator is modeled as the difference between the target joint position, denoted as \(\mathbf{\theta}_{\text{target}}\), and the nominal joint position, \(\mathbf{\theta}_0\). To mitigate the instability in the network's output, we introduce a scaling factor \(k \leq 1\) that is applied to the policy output \(\mathbf{a}_{t}\). Consequently, the final target joint positions are given by the equation:  
\(\mathbf{\theta}_{\text{target}} = \mathbf{\theta}_0 + k \mathbf{a}_{t}\).
The action space \(\mathcal{A}\) is defined by the number of actuators in the system. For instance, in the case of quadrupedal robots such as the Unitree A1, which is equipped with 12 actuators (3 on each leg), the dimension of the action space is 12.

%% file: section/4_method.tex
\subsection{Framework Overview}

Our framework is shown in Fig.~\ref{fig:framework}.
It is a general approach that can be incorporated into most of the previous RL-based Locomotion methods such as RMA~\cite{kumar2021rma}, Dreamwaq~\cite{nahrendra2023dreamwaq}, HIMLoco~\cite{long2024him}, etc., and here we use HIMLoco as the backbone algorithm.
It consists of two parts: Motion Reference Learning and Policy Learning with Imagined Transition.
Section~\ref{sec4_1} discusses how to perform Motion Reference Learning. It aims to obtain the optimal gait from a fixed simulation environment and compress the optimal gait into an ideal policy and a dynamics model.
Section~\ref{sec4_2} presents how to use motion reference to guide policy learning.

\subsection{Motion Reference Learning in Fixed Dynamics}
\label{sec4_1}
\subsubsection{Ideal Policy}
In the process of reinforcement learning to train robotic policies, we found that the wider ranges of dynamic randomization, the lower returns that can be obtained when the training converges.
To obtain an optimal reference, we train a policy that is optimal in the unperturbed simulation environment in a fixed dynamic (i.e., no dynamic randomization setting) environment, which is called the ideal policy $ \pi_{ideal}(\mathbf{a}^r_t|\mathbf{o}_{t:t-H}) $ in the paper.

\subsubsection{Dynamics Model}
\begin{figure}[h]
    \centering
    \includegraphics[width=0.8\linewidth]{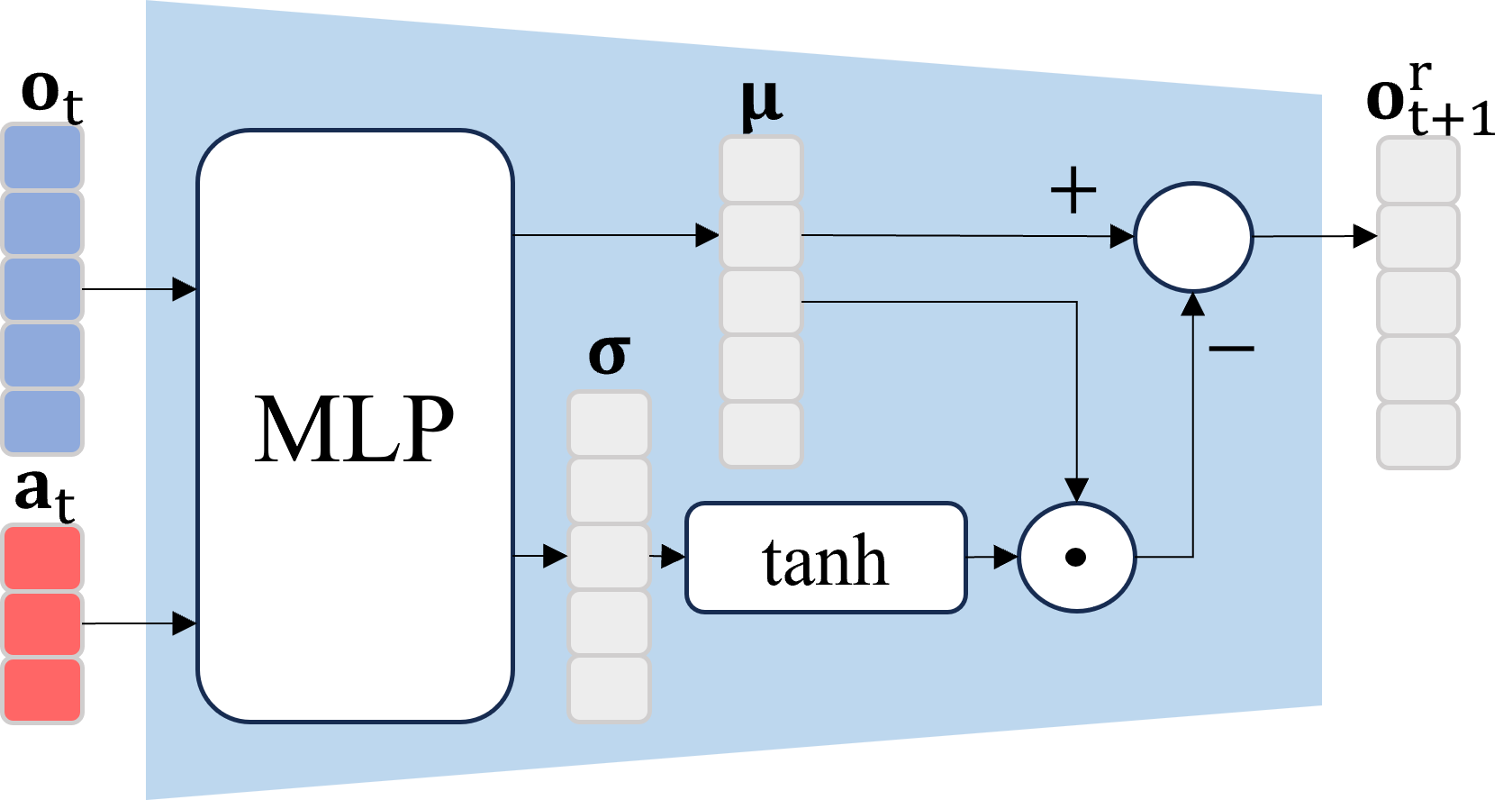}
    \caption{\textbf{Architecture of Dynamics Model.} An adjustment mechanism is added to the output of the dynamics model.}
    \label{fig:model}
\end{figure}
To obtain reference observations, we train a dynamics model under the non-perturbative simulation environment.
The output of the dynamics model includes $\mu$ and  $\sigma$ in eq.~\ref{eq:output}, which means the mean and standard deviation of the next observation estimation.
We use the dynamics model to output $\mu$ and $\sigma$, which parameterize a normal distribution. The training objective is to maximize the likelihood of the true next observation $\mathbf{o}_{t+1}$ under this distribution.

\begin{equation}
    \mu, \sigma = d_{\theta}(\cdot | \mathbf{o}_t, \mathbf{a}_t)
    \label{eq:output}
\end{equation}
\begin{equation}
    \mathcal{L}_{MLE} = -\mathbb{E}[\log p(\mathbf{o}_{t+1}|\mu,\sigma)]
    \label{eq:mle}
\end{equation}

\begin{figure}[h]
    \centering
    \includegraphics[width=1\linewidth]{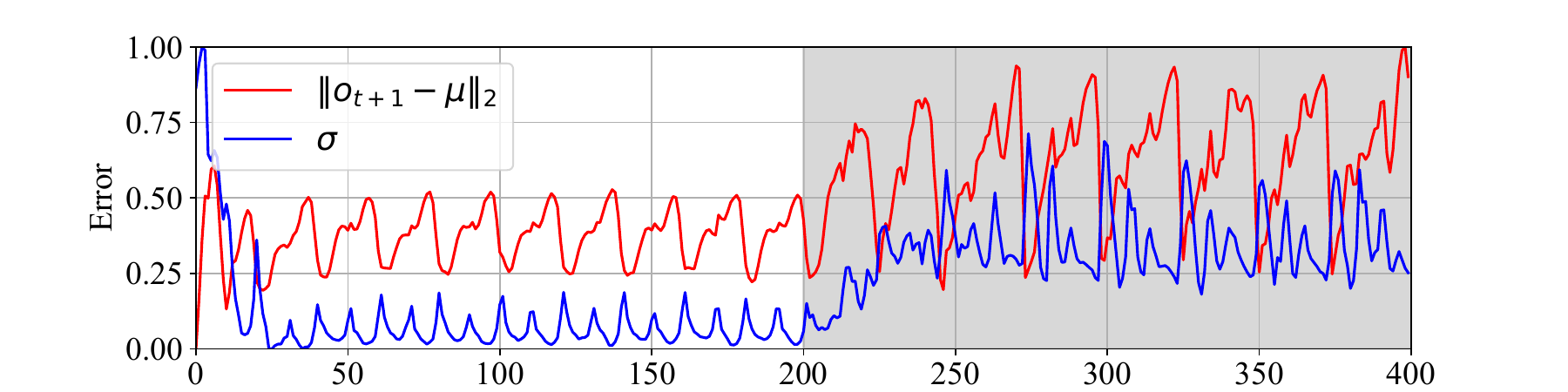}
    \caption{Correlation between standard deviation estimate and actual observation bias. Unseen disturbance is applied to the robot after 200 steps.}
    \label{fig:stdev}
\end{figure}

In most model-based methods, model error is an unavoidable problem.
We measured prediction error  \( \|{\mathbf{o}}_{t+1} - \mu\|_2 \) while applying the unseen disturbance in the robot's body or not, and compared it to \( \sigma \).
The results in Fig.~\ref{fig:stdev} show a strong correlation between actual prediction error and $\sigma$ given by the learned model.

To prevent training and deployment collisions caused by model inaccuracies, we developed an adjustment mechanism to adjust the prediction of the next observation in the following robust policy learning stage (see Sec.~\ref{sec4_2}) and deployment.
According to the general law of supervised learning, the larger the standard deviation $\sigma$ the less reliable the prediction $\mu$ of the dynamics model.
Thus, when the variance is large, we should minimize the effect of the observation reference on the subsequent network. 
Inspired by~\cite{nahrendra2023dreamwaq, nahrendra2024obstacle}, we multiply the weight $(1- norm(\sigma)) \in [0, 1]$ dot by $\mu$ with the following equation~\ref{eq:adjust}.
Experiments in~\ref{fig:payload} showed that this design significantly improved the robot's performance in the unseen region.

\begin{equation}
    \mathbf{o}^{r}_{t+1} = (1- norm(\sigma))\cdot\mu
    \label{eq:adjust}
\end{equation}

\subsection{Policy Learning with Imagined Transition}
\label{sec4_2}
\begin{figure}[h]
    \centering
    \includegraphics[width=0.9\linewidth]{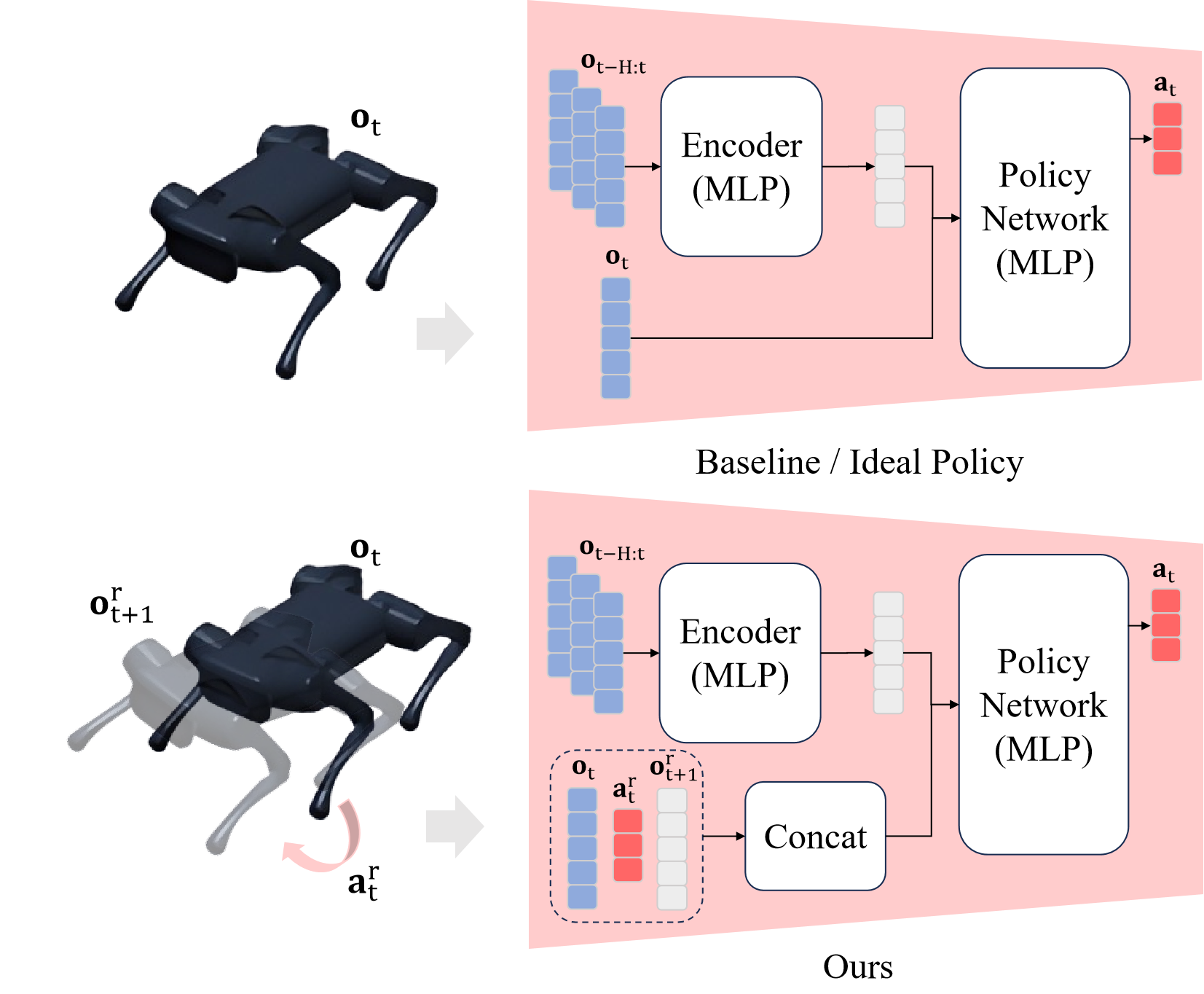}
    \caption{\textbf{Architecture of Policy Network.} The top is the baseline architecture, and the bottom is the architecture of the policy network with imagined transition.}
    \label{fig:actor}
\end{figure}

In short, the only difference in our policy network is that we are adding the input of action and observation reference, which represents motion reference.
The combination of $\mathbf{o}$, $\mathbf{a}^{r}_{t}$, and $\mathbf{o}^{r}_{t+1}$ forms a complete state transition, which is our imagined transition.
As shown in the bottom of Fig.\ref{fig:actor}, we concatenate reference action $\mathbf{a}^{r}_{t}$ and imagined next observation $\mathbf{o}^{r}_{t+1}$ from the motion reference with proprioceptive observations $\mathbf{o}_{t}$, whereas the baseline, depicted on the top, follows the encoder-policy network architecture used in\cite{nahrendra2023dreamwaq, long2024him}.

\begin{equation}
    \mathbf{a}_t \sim \pi(\cdot | \mathbf{o}_{t-H:t})
\end{equation}
\begin{equation}
    \mathbf{a}_t \sim \pi(\cdot | \mathbf{o}_{t-H:t}, \mathbf{a}^{r}_{t}, \mathbf{o}^{r}_{t+1})
\end{equation}

Intuitively, solving in the solution space close to the motion prior should be more efficient than searching for the optimal solution directly in the huge solution space.
This simple change has resulted in a significant increase in the overall performance of policy learning.

%% file: section/5_experiment.tex
\subsection{Experimental Setup}
\paragraph{Compared Methods}
\begin{figure}[h]
    \centering
    \includegraphics[width=1\linewidth]{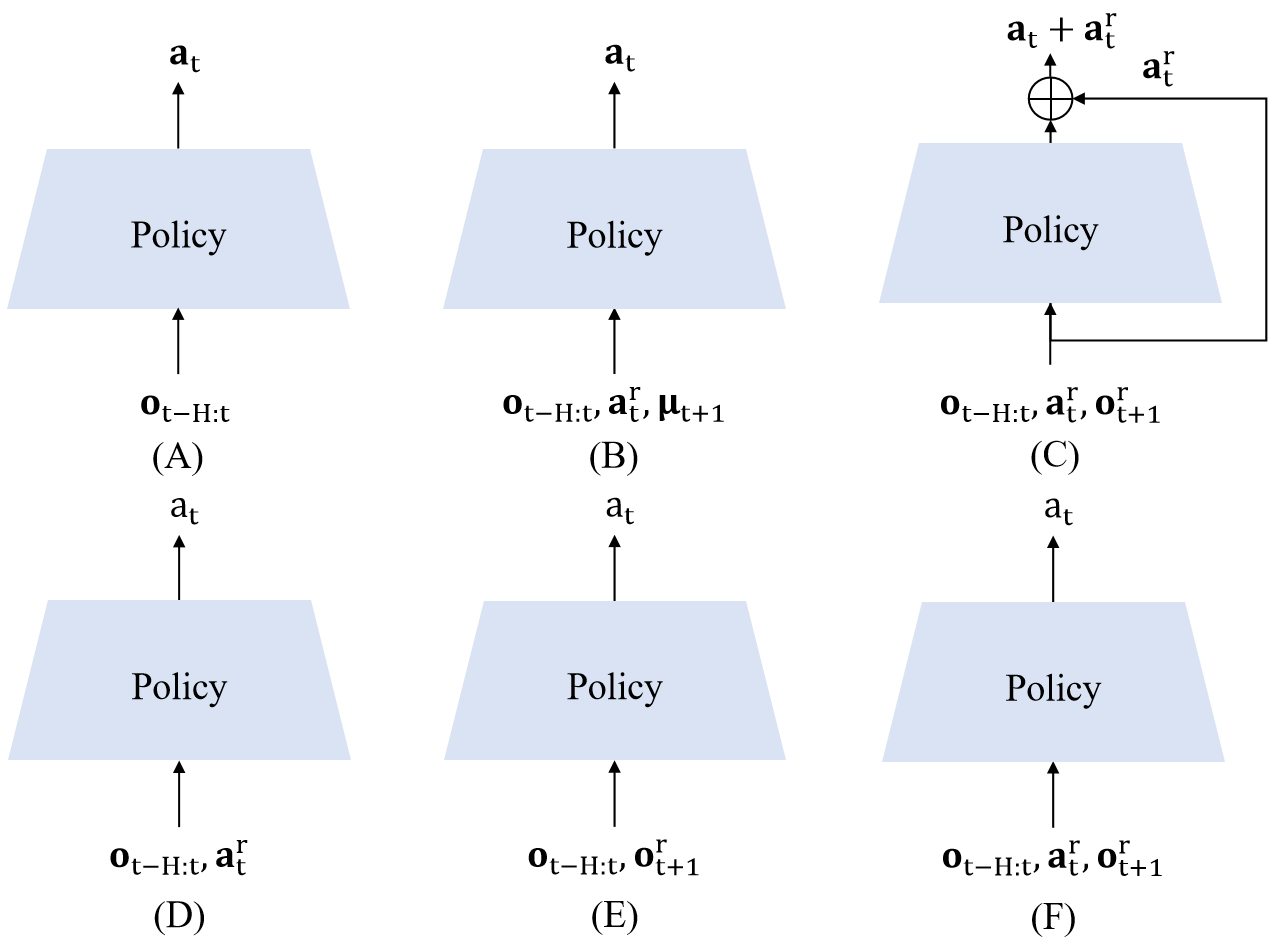}
    \caption{Compared Methods.}
    \label{fig:baseline}
\end{figure}

To fairly compare our method with others, we set up five variants as Figure~\ref{fig:baseline}.

(A) Baseline: Train with current and history observations, equivalent here to~\cite{long2024him};
(B) Ours w/o Adjust: Ours without observation reference adjustment, the mean value output from the dynamics model is used directly as the observation reference;
(C) Ours w Residual: Action reference is summed with the actor output as action output, similar to residual policy in~\cite{silver2018residual, yuan2024policy};
(D) Ours w/o Action Reference: Action reference input is set to zero tensor;
(E) Ours w/o State Reference: State reference input is set to zero tensor;
(F) Ours;

\paragraph{Simulation Setup} We use the PPO algorithm with 4096 parallel environments and a rollout length of 100 time steps in Isaac Gym~\cite{rudin2022learning}. The training process takes 4096 parallel environments for 2000 iterations in the NVIDIA A800 GPU. In Section \ref{sec5_2_2}, we trained longer iterations to observe learning curves.

\subsection{Evaluation on Simulator}
\subsubsection{Command Tracking}
We evaluate the tracking errors in linear velocity and angular velocity per second for each algorithm under various terrains.
Tracking errors are calculated by
${\|v_{x,y} - v_{x,y}^{\text{target}}\|^2}$ and ${\|\omega_{\text{yaw}} - \omega_{\text{yaw}}^{\text{target}}\|^2}$ 
respectively.
The results in Table~\ref{tab:tracking_error} show that our method has a lower velocity tracking error.

\begin{table}[thb]
%\footnotesize
    \centering
    \caption{Average Tracking Error in Simulator over 1000 Trials.}
    \label{tab:tracking_error}
    \scalebox{1.2}{
    \begin{tabular}{llllll}
        \hline
        \makecell[l]{Terrain Types} & \makecell[l]{Velocity Types} & Baseline & Ours \\
        \hline
        \multirow{2}{*}{\makecell[l]{Smooth Slopes}} 
        & Linear & 0.170 & \textbf{0.120} \\
        & Angular & 0.101 & \textbf{0.099} \\
        \hline
        \multirow{2}{*}{\makecell[l]{Rough Slopes}} 
        & Linear & 0.387 & \textbf{0.162} \\
        & Angular & 0.164 & \textbf{0.130} \\
        \hline
        \multirow{2}{*}{\makecell[l]{Stairs}} 
        & Linear & 2.371 & \textbf{0.992} \\
        & Angular & \textbf{0.515} & 0.552 \\
        \hline
        \multirow{2}{*}{\makecell[l]{Discrete}} 
        & Linear & 1.955 & \textbf{0.120} \\
        & Angular & 0.220 & \textbf{0.099} \\
        \hline
    \end{tabular}
    }
    \vspace{-0.2in}
\end{table}

\subsubsection{Learning Curve}
\label{sec5_2_2}
\begin{figure}[h]
    \centering
    \includegraphics[width=1\linewidth]{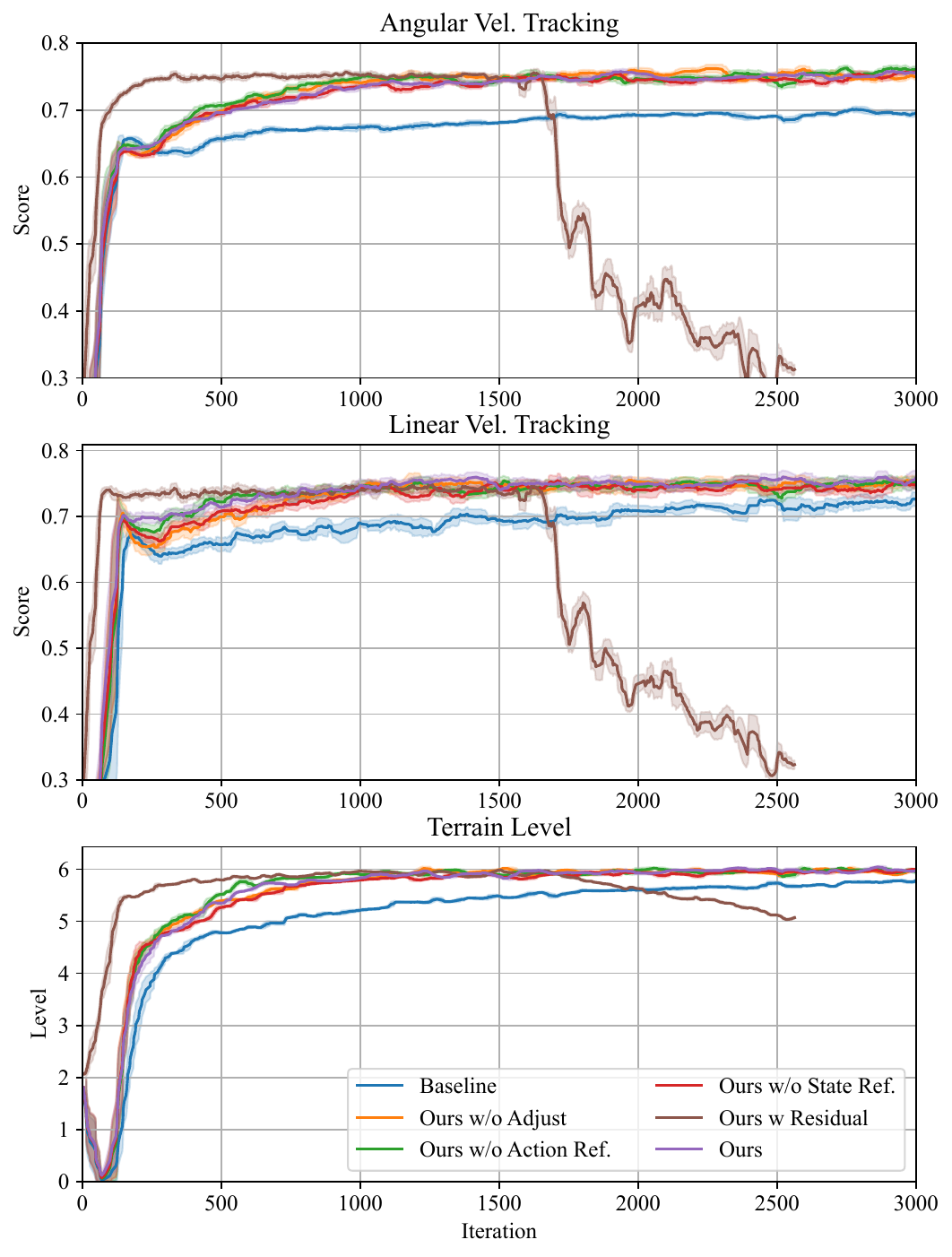}
    \caption{\textbf{Ablation studies with learning curves.} From top to bottom: (1) normalized linear velocity tracking score, (2) normalized angular velocity tracking score, and (3) maximum reachable terrain level in training.}
    \label{fig:curve}
\end{figure}
We recorded the growth of velocity tracking score and terrain passing ability for different algorithms during training.
The normalized angular velocity tracking score and the normalized linear velocity tracking score are calculated by $\text{exp}(-\frac{\|\omega_{\text{yaw}} - \omega_{\text{yaw}}^{\text{target}}\|^2}{0.25})$ 
and $\text{exp}(-\frac{\|v_{x,y} - v_{x,y}^{\text{target}}\|^2}{0.25})$
respectively.
The terrain is categorized into 9 difficulty levels based on the curriculum setting, the same as~\cite{long2024him}.

Observing from the learning curves in Figure~\ref{fig:curve}, our method consistently maintains an advantage over the baseline in terms of tracking error and terrain traversal ability throughout the training process.
Notably, after the introduction of action residual, the method crashed in the middle of training. Based on previous analysis, there is a distribution shift between reference learning and policy learning. 
Although residual concatenation of reference action of imagined transitions improves the speed of policy learning early in training, the imagined transitions are unreliable when encountering the OOD state, leading to potential policy training breakdowns.
Our adjustment mechanism of the dynamics model aims to mitigate the adverse effects of this OOD issue.

\subsubsection{Payload Shift}
\begin{figure}[h]
    \centering
    \includegraphics[width=1\linewidth]{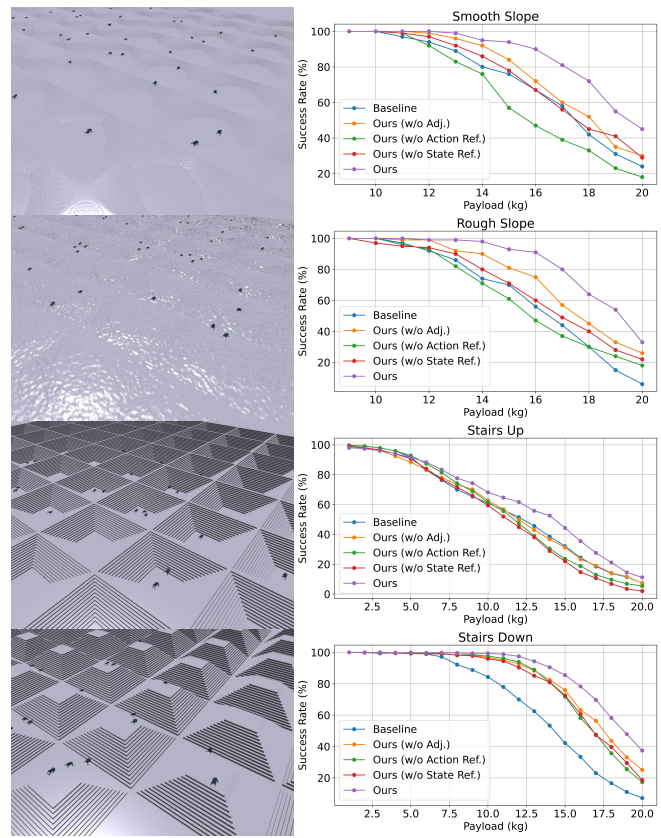}
    \caption{\textbf{Success rate of different algorithms under various payloads.} The success rate for payloads from 1kg to 9kg on both smooth and rough slopes is 100\%.}
    \label{fig:payload}
\end{figure}

\begin{figure*}[htb]
    \centering
    \includegraphics[width=1\linewidth]{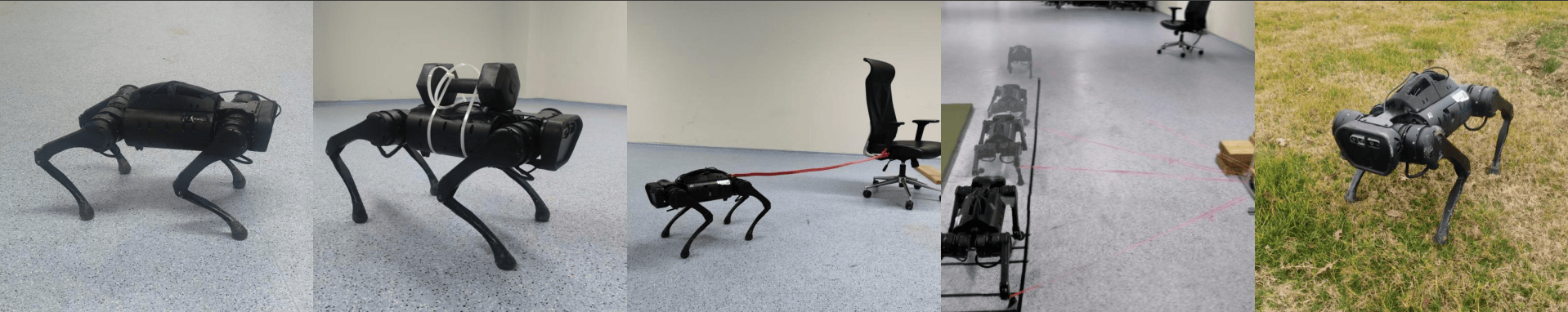}
    \caption{\textbf{Deployment on Real World.} Five scenarios in order from left to right: (1) Plane - flat surface walking, (2) Payload - carrying 3 Kg additional weight, (3) Disturbance-1 - external force applied from the back, (4) Disturbance-2 - external disturbance applied to the right-back leg, and (5) Lawn - walking on uneven grassy terrain.}
    \label{fig:realworld}
\end{figure*}

To verify the performance of our approach against the unknown heavy payload, we tested the success rate of the A1 robot in executing different payloads on four different terrains: smooth slopes, rough slopes, stairs up, and stairs down with varying levels of difficulty. 
The robot was required to walk forward with the command [1, 0, 0](m/s) in all four terrains for 200 steps. The range of payload gradually increases from 1kg to 20kg. For smooth and rough slopes, we execute 100 times with different seeds, and since the terrain of stairs is more complicated, we execute 1000 times to calculate the success rate, and the result is shown in Figure~\ref{fig:payload}.

On all test terrains, our method survived significantly better than the other compared methods when faced with unseen heavy loads.
In addition, we can find that although the changes in state adjustment and action reference input do not show significant differences in the learning curve and the evaluation within the distribution, they affect the performance of the method in unseen payloads.

\subsubsection{Adding push to robots}
\begin{figure}[h]
    \centering
    \includegraphics[width=1\linewidth]{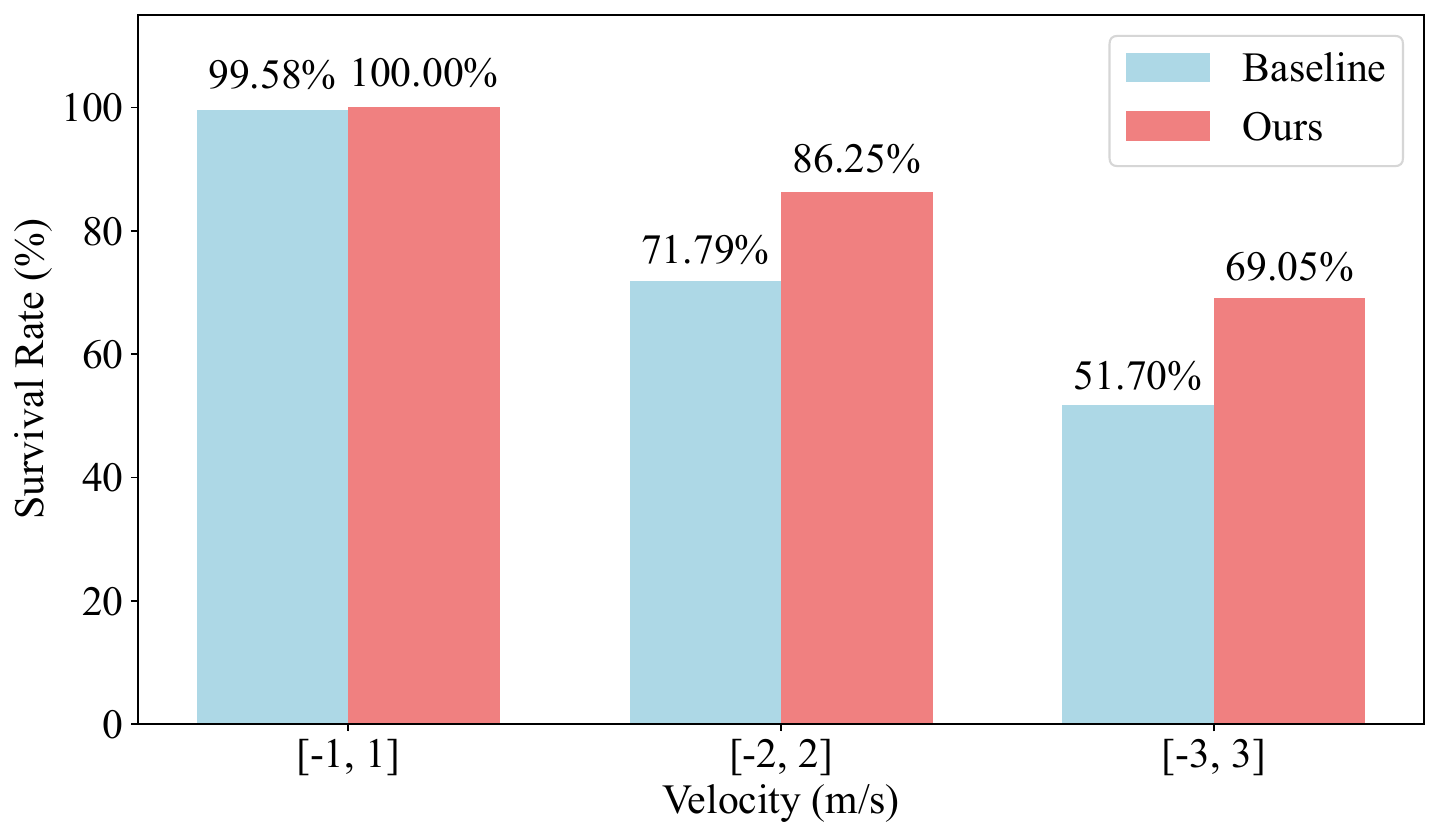}
    \caption{\textbf{Comparison of survival rate under external pushing.} The experimental results are categorized into three classes based on the magnitude of the external push velocity. A higher survival rate indicates better robustness.}
    \label{fig:push_bar}
\end{figure}

Pushing a robot with external force is a widely used method for evaluating the robustness of its locomotion. 
In this experiment, the push force was applied by adding a random velocity to the quadruped’s base along the \(xy\)-plane of the world frame. 
The robot was instructed to walk forward with a command of \([1, 0, 0] \, \text{m/s}\), while simultaneously responding to the unpredicted random push. 
The magnitude of the added velocity determines the intensity of the push, with larger magnitudes corresponding to more forceful pushes. 
To assess the robustness of our approach in comparison to the baseline, pushes were randomly sampled 2000 times within a velocity range of \([-3, 3] \, \text{m/s}\). 
The results in Figure~\ref{fig:push_bar} illustrate that our method outperforms the baseline in terms of fall resistance when subjected to external pushes.

% \begin{table}[h]
% %\footnotesize
% \tabcolsep=2pt
%     \centering
%     \caption{Survive Rate under Pushing at Different Velocities.(\%(survives/total))}
%     \label{tab:vel_success_rate}
%     \scalebox{1.2}{
%     \begin{tabular}{cccc}
%         \hline
%         Velocity (m/s) & Baseline &
%         Ours & Improvement(\%) \\
%         \hline
%         $[-1, 1]$ & \makecell[c]{99.58\\(237/238)} & \makecell[c]{100\\(238/238)} & 0.42 $\uparrow$ \\
%         $[-2, 2]$ & \makecell[c]{71.79\\(616/858)} & \makecell[c]{86.25\\(740/858)} & 14.46 $\uparrow$ \\
%         $[-3, 3]$ & \makecell[c]{51.70\\(1034/2000)} & \makecell[c]{69.05\\(1381/2000)} & 17.35 $\uparrow$ \\
%         \hline
%     \end{tabular}
%     }
% \end{table}

\begin{figure}[h]
    \centering
    \includegraphics[width=0.8\linewidth]{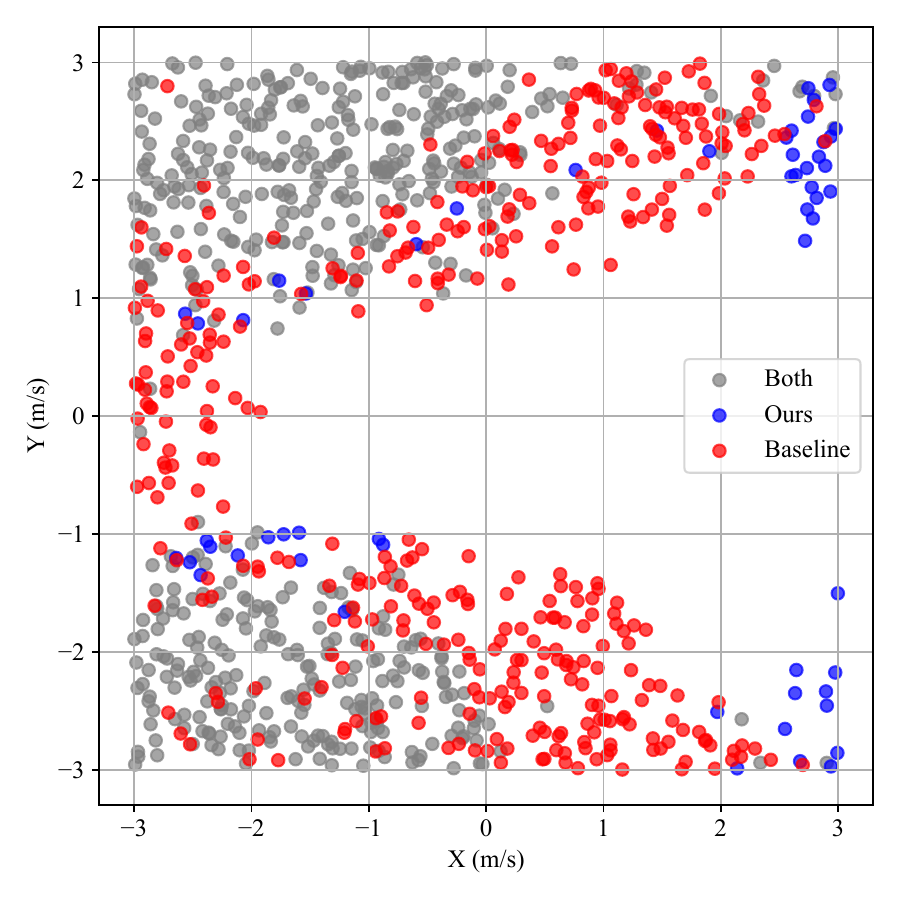}
    \caption{\textbf{Fall resistant visualization with external push.} `Both' represents cases where both ours and baseline fall. `Ours' and `Baseline' represent cases where each falls but the other survives.}
    \label{fig:push}
\end{figure}

Figure~\ref{fig:push} provides a detailed visualization of the standing versus falling status under various push magnitudes and directions for different algorithms.
Given the command \([1, 0, 0] \, \text{m/s}\), it can be observed that Figure~\ref{fig:push} exhibits near-symmetry along the \(x\)-axis, but not along the \(y\)-axis. All methods include push perturbations within the range \([-1, 1] \, \text{m/s}\) during the robust training phase, and consequently, they perform well within this range. However, as the perturbation range increases, the advantage of our approach becomes increasingly pronounced. Our method outperforms the baseline by surviving more than 300 additional times in 2000 randomized tests.

\subsection{Evaluation on Real World}
\begin{figure}[h]
    \centering
    \includegraphics[width=1\linewidth]{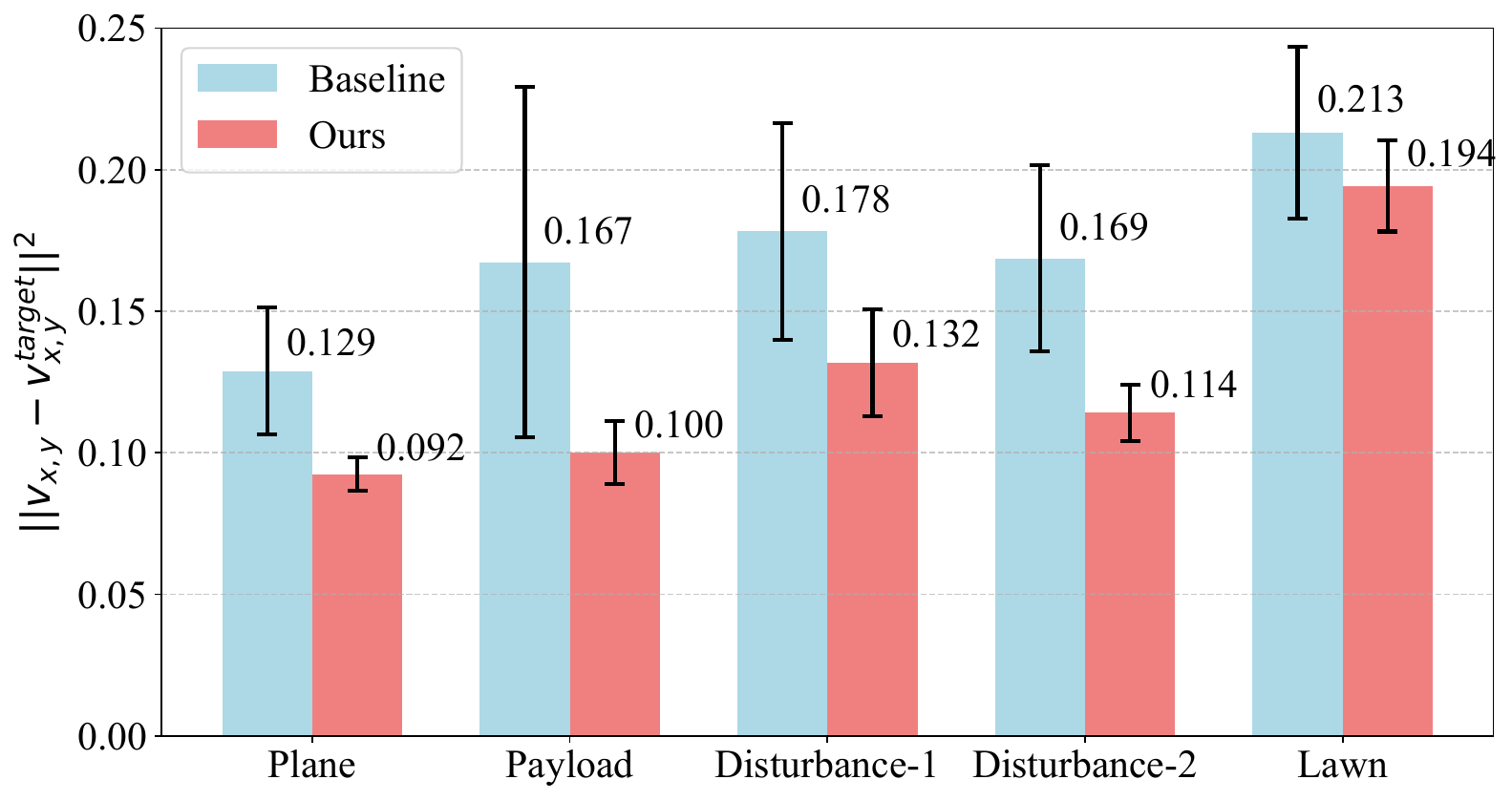}
    \caption{\textbf{Comparison of command tracking performance in real-world scenarios.} The results are averaged over ten trials. Error bars represent standard error. A lower tracking error indicates better deployed performance.}
    \label{fig:realworld_bar}
\end{figure}

To test the effectiveness of policy deployment in the real world, we designed a variety of challenging scenarios as shown in Figure~\ref{fig:realworld}.
We deployed the policy for training 2000 iterations on the UniTree A1 robot, with the PD controller's parameters set to $Kp=40.0$ and $Kd=1.0$.
The robot was instructed to walk forward with a command of \([1, 0, 0] \, \text{m/s}\) in five scenarios.
We measured the tracking errors \(||v_{x,y} - v^{target}_{x,y}||^2\) as the performance metric to evaluate the robot's ability to track the linear velocity command, where the body velocity \(v_{x,y}\) is collected by the robot's IMU.
The results in Figure~\ref{fig:realworld_bar} show that our method has a lower tracking error in five scenarios compared to the baseline.

\subsection{Implementation Details}
\paragraph{Reward Design}
The design of reward functions follows previous research\cite{nahrendra2023dreamwaq} and is shown in Table~\ref{tab:reward}.
\begin{table}[h]
% \footnotesize
% \tabcolsep=2pt
    \centering
    \caption{Reward functions.}
    \scalebox{1.05}{
    \begin{tabular}{lll}
    \toprule
    Reward & Equation $\left(r_i\right)$ & Weight $\left(w_i\right)$ \\
    \midrule
    \makecell[l]{Linear velocity\\tracking} & $\exp \left\{- \frac{\|\mathbf{v}_{x y}^{\text{cmd}}-\mathbf{v}_{x y}\|_2^2}{0.25}  \right\}$ & 1.0 \\
    \makecell[l]{Angular velocity\\tracking} & $\exp \left\{- \frac{\left(\omega_{\text {yaw}}^{\text{cmd}}-\omega_{\text{yaw}}\right)^2}{0.25} \right\}$ & 0.5 \\
    Lin. velocity $(z)$ & $v_z^2$ & -2.0 \\
    Ang. velocity $(x y)$ & $\|\boldsymbol{\omega}_{x y}\|_2^2$ & -0.05 \\
    Orientation & $\|\mathbf{g}\|_2^2$ & -0.2 \\
    Joint accelerations & $\|\ddot{\boldsymbol{\theta}}\|^2$ & $-2.5 \times 10^{-7}$ \\
    Joint power & $|\boldsymbol{\tau} \| \dot{\boldsymbol{\theta}}|^{T}$ & $-2 \times 10^{-5}$ \\
    Body height & $\left(h^{\text {target}}-h\right)^2$ & -1.0 \\
    Foot clearance & $\sum_{i=0}^{3}\left(p_{z}^{\text {target}}-p_{z}^{i}\right)^2 \cdot v_{xy}^{i}$ & -0.01 \\
    Action rate & $\|\mathbf{a}_t-\mathbf{a}_{t-1}\|_2^2$ & -0.01 \\
    Smoothness & $\|\mathbf{a}_t-2 \mathbf{a}_{t-1}+\mathbf{a}_{t-2}\|_2^2$ & -0.01 \\
    \bottomrule
    \end{tabular}
    }
    \label{tab:reward}
\end{table}

Where $h^{\text{target}}$ is the desired base height corresponding to ground, $p_z^{\text{target}}$ and $p_z^i$ are the desired feet position and real feet position in the z-axis of robots' frame and $v_{xy}^i$ is the feet velocity in xy-plane of robot's frame.

\paragraph{Dynamics Randomization} We randomize the mass of the robot body and links, the center of mass (CoM) of the robot, the payload applied to the body of the robot, the ground friction and restitution coefficients, the motor strength, the joint-level PD gains, the system delay, the external force, and the initial joint positions during policy Learning with imagined transition. The randomization ranges for each parameter are detailed in Table~\ref{tab:random}.

\begin{table}[h]
% \footnotesize
    \centering
    \caption{Domain Randomization.}
    \scalebox{1.05}{
    \begin{tabular}{lll}
    \toprule 
    Parameters & Range & Unit \\
    \midrule
    CoM & {$[-0.05,0.05]$} & $\mathrm{m}$ \\
    Payload Mass & {$[-1,2]$} & $\mathrm{Kg}$ \\
    Ground Friction & {$[0.2,1.25]$} & - \\
    Motor Strength & {$[0.9,1.1] \times$ motor torque } & $\mathrm{Nm}$ \\
    Joint $K_p$ & {$[0.9,1.1] \times$ 40.0} & Nm/rad \\
    Joint $K_d$ & {$[0.9,1.1] \times$ 1.0} & Nms/rad \\
    Initial Joint Positions & {$[0.5,1.5] \times$ nominal value } & $\mathrm{rad}$ \\
    System Delay & {$[0,15]$ } & $\mathrm{ms}$ \\
    External Force & {$[-30, 30]$} & $\mathrm{N}$ \\
    \bottomrule
    \end{tabular}
    }
    \label{tab:random}
\end{table}

\paragraph{Training Curriculum}
To facilitate robot learning, we adopt a terrain curriculum and command curriculum inspired by the approaches of \cite{rudin2022learning}, \cite{wu2023learning}, and \cite{long2024him}. 
Specifically, we construct a height field map comprising 200 distinct terrains organized in a $20 \times 10$ grid. Within each row, terrains are arranged in increasing order of difficulty, and each grid cell measures $10 \times 10 \, m^{2}$. 
Initially, robots are placed on the simplest terrain types. As the robot's performance improves, the terrain difficulty level increases once the robot achieves at least 80\% of the linear tracking reward. Conversely, the difficulty decreases if the robot cannot traverse half of the terrain within a single episode.
For terrains involving stairs or discrete obstacles, we sample longitudinal and lateral linear velocity commands from the range $[-1.0, 1.0]$ m/s and the horizontal angular velocity from $[-2.0, 2.0]$ rad/s. For slopes and rough terrains, the range of longitudinal linear velocity is extended to $[-3.0, 3.0]$ m/s, and the range for horizontal angular velocity is expanded to $[-3.0, 3.0]$ rad/s. All commands are sampled independently for each robot from the specified ranges, with sampling occurring every 25-time steps.

%% file: section/6_conclusion.tex
In this paper, we propose LIT, a two-stage framework that aims to mitigate the trade-off between robustness and optimality in RL-based locomotion. 
By introducing an imagined transition to policy learning, LIT enables policies trained with domain randomization to be guided by the desired gait in the simulator. 
Extensive experiments demonstrate significant improvements in velocity tracking under both simulators and real-world environments. 
Inspired by reference signals in classical control methods, LIT provides the policy with reference actions and imagined next observations, supplies richer contextual information, enabling it to anticipate better and adapt to dynamic variations introduced by domain randomization.
In future work, we will continue to investigate robotic scenarios exhibiting inherent trade-offs between precision and robustness, while exploring generalized implementations of this paradigm across broader application domains.

%% file: root.bbl
\begin{thebibliography}{10}
\providecommand{\url}[1]{#1}
\csname url@rmstyle\endcsname
\providecommand{\newblock}{\relax}
\providecommand{\bibinfo}[2]{#2}
\providecommand\BIBentrySTDinterwordspacing{\spaceskip=0pt\relax}
\providecommand\BIBentryALTinterwordstretchfactor{4}
\providecommand\BIBentryALTinterwordspacing{\spaceskip=\fontdimen2\font plus
\BIBentryALTinterwordstretchfactor\fontdimen3\font minus \fontdimen4\font\relax}
\providecommand\BIBforeignlanguage[2]{{%
\expandafter\ifx\csname l@#1\endcsname\relax
\typeout{** WARNING: IEEEtran.bst: No hyphenation pattern has been}%
\typeout{** loaded for the language `#1'. Using the pattern for}%
\typeout{** the default language instead.}%
\else
\language=\csname l@#1\endcsname
\fi
#2}}

\bibitem{xie2021dynamics}
Z.~Xie, X.~Da, M.~van~de Panne, B.~Babich, and A.~Garg, ``Dynamics randomization revisited: A case study for quadrupedal locomotion,'' in \emph{Proc. IEEE International Conference on Robotics and Automation (ICRA)}, 2021, pp. 4955--4961.

\bibitem{tan2018sim}
J.~Tan, T.~Zhang, E.~Coumans, A.~Iscen, Y.~Bai, D.~Hafner, S.~Bohez, and V.~Vanhoucke, ``Sim-to-real: Learning agile locomotion for quadruped robots,'' \emph{arXiv preprint arXiv:1804.10332}, 2018.

\bibitem{kumar2021rma}
A.~Kumar, Z.~Fu, D.~Pathak, and J.~Malik, ``Rma: Rapid motor adaptation for legged robots,'' in \emph{Robotics: Science and Systems}, 2021.

\bibitem{nahrendra2023dreamwaq}
I.~M.~A. Nahrendra, B.~Yu, and H.~Myung, ``Dreamwaq: Learning robust quadrupedal locomotion with implicit terrain imagination via deep reinforcement learning,'' in \emph{IEEE International Conference on Robotics and Automation (ICRA)}, 2023.

\bibitem{margolis2023walk}
G.~B. Margolis and P.~Agrawal, ``Walk these ways: Tuning robot control for generalization with multiplicity of behavior,'' in \emph{Conference on Robot Learning (CoRL)}, 2023.

\bibitem{long2024him}
J.~Long, Z.~Wang, Q.~Li, L.~Cao, J.~Gao, and J.~Pang, ``Hybrid internal model: Learning agile legged locomotion with simulated robot response,'' in \emph{The Twelfth International Conference on Learning Representations}, 2024.

\bibitem{tobin2017domain}
J.~Tobin, R.~Fong, A.~Ray, J.~Schneider, W.~Zaremba, and P.~Abbeel, ``Domain randomization for transferring deep neural networks from simulation to the real world,'' in \emph{2017 IEEE/RSJ international conference on intelligent robots and systems (IROS)}.\hskip 1em plus 0.5em minus 0.4em\relax IEEE, 2017, pp. 23--30.

\bibitem{pmlr-v100-mehta20a}
\BIBentryALTinterwordspacing
B.~Mehta, M.~Diaz, F.~Golemo, C.~J. Pal, and L.~Paull, ``Active domain randomization,'' in \emph{Proceedings of the Conference on Robot Learning}, ser. Proceedings of Machine Learning Research, L.~P. Kaelbling, D.~Kragic, and K.~Sugiura, Eds., vol. 100.\hskip 1em plus 0.5em minus 0.4em\relax PMLR, 30 Oct--01 Nov 2020, pp. 1162--1176. [Online]. Available: \url{https://proceedings.mlr.press/v100/mehta20a.html}
\BIBentrySTDinterwordspacing

\bibitem{tiboni2023doraemon}
G.~Tiboni, P.~Klink, J.~Peters, T.~Tommasi, C.~D'Eramo, and G.~Chalvatzaki, ``Domain randomization via entropy maximization,'' 2023.

\bibitem{todorov2012mujoco}
E.~Todorov, T.~Erez, and Y.~Tassa, ``Mujoco: A physics engine for model-based control,'' in \emph{2012 IEEE/RSJ international conference on intelligent robots and systems}.\hskip 1em plus 0.5em minus 0.4em\relax IEEE, 2012, pp. 5026--5033.

\bibitem{makoviychuk2021isaac}
V.~Makoviychuk, L.~Wawrzyniak, Y.~Guo, M.~Lu, K.~Storey, M.~Macklin, D.~Hoeller, N.~Rudin, A.~Allshire, A.~Handa, \emph{et~al.}, ``Isaac gym: High performance gpu-based physics simulation for robot learning,'' \emph{Advances in neural information processing systems}, 2021.

\bibitem{rudin2022learningwalkminutesusing}
\BIBentryALTinterwordspacing
N.~Rudin, D.~Hoeller, P.~Reist, and M.~Hutter, ``Learning to walk in minutes using massively parallel deep reinforcement learning,'' 2022. [Online]. Available: \url{https://arxiv.org/abs/2109.11978}
\BIBentrySTDinterwordspacing

\bibitem{doi:10.1126/scirobotics.abc5986}
\BIBentryALTinterwordspacing
J.~Lee, J.~Hwangbo, L.~Wellhausen, V.~Koltun, and M.~Hutter, ``Learning quadrupedal locomotion over challenging terrain,'' \emph{Science Robotics}, vol.~5, no.~47, p. eabc5986, 2020. [Online]. Available: \url{https://www.science.org/doi/abs/10.1126/scirobotics.abc5986}
\BIBentrySTDinterwordspacing

\bibitem{pmlr-v205-agarwal23a}
\BIBentryALTinterwordspacing
A.~Agarwal, A.~Kumar, J.~Malik, and D.~Pathak, ``Legged locomotion in challenging terrains using egocentric vision,'' in \emph{Proceedings of The 6th Conference on Robot Learning}, ser. Proceedings of Machine Learning Research, K.~Liu, D.~Kulic, and J.~Ichnowski, Eds., vol. 205.\hskip 1em plus 0.5em minus 0.4em\relax PMLR, 14--18 Dec 2023, pp. 403--415. [Online]. Available: \url{https://proceedings.mlr.press/v205/agarwal23a.html}
\BIBentrySTDinterwordspacing

\bibitem{Cheng2024parkour}
X.~Cheng, K.~Shi, A.~Agarwal, and D.~Pathak, ``Extreme parkour with legged robots,'' in \emph{2024 IEEE International Conference on Robotics and Automation (ICRA)}, 2024, pp. 11\,443--11\,450.

\bibitem{kumar2022adapting}
A.~Kumar, Z.~Li, J.~Zeng, D.~Pathak, K.~Sreenath, and J.~Malik, ``Adapting rapid motor adaptation for bipedal robots,'' in \emph{2022 IEEE/RSJ International Conference on Intelligent Robots and Systems (IROS)}.\hskip 1em plus 0.5em minus 0.4em\relax IEEE, 2022, pp. 1161--1168.

\bibitem{ji2022concurrent}
G.~Ji, J.~Mun, H.~Kim, and J.~Hwangbo, ``Concurrent training of a control policy and a state estimator for dynamic and robust legged locomotion,'' \emph{IEEE Robotics and Automation Letters}, vol.~7, no.~2, pp. 4630--4637, 2022.

\bibitem{higgins2016beta}
I.~Higgins, L.~Matthey, A.~Pal, C.~Burgess, X.~Glorot, M.~Botvinick, S.~Mohamed, and A.~Lerchner, ``$\beta$ -- {VAE}: Learning basic visual concepts with a constrained variational framework,'' in \emph{Proc. International Conference on Learning Representations (ICLR)}, 2017.

\bibitem{lyu2023composite}
S.~Lyu, H.~Zhao, and D.~Wang, ``A composite control strategy for quadruped robot by integrating reinforcement learning and model-based control,'' in \emph{2023 IEEE/RSJ International Conference on Intelligent Robots and Systems (IROS)}.\hskip 1em plus 0.5em minus 0.4em\relax IEEE, 2023, pp. 751--758.

\bibitem{lyu2024rl2ac}
S.~Lyu, X.~Lang, H.~Zhao, H.~Zhang, P.~Ding, and D.~Wang, ``Rl2ac: Reinforcement learning-based rapid online adaptive control for legged robot robust locomotion,'' in \emph{Proceedings of the Robotics: Science and Systems}, 2024.

\bibitem{kim2019highly}
D.~Kim, J.~Di~Carlo, B.~Katz, G.~Bledt, and S.~Kim, ``Highly dynamic quadruped locomotion via whole-body impulse control and model predictive control,'' \emph{arXiv preprint arXiv:1909.06586}, 2019.

\bibitem{ioannou1996robust}
P.~A. Ioannou and J.~Sun, \emph{Robust adaptive control}.\hskip 1em plus 0.5em minus 0.4em\relax PTR Prentice-Hall Upper Saddle River, NJ, 1996, vol.~1.

\bibitem{hafner2019dream}
D.~Hafner, T.~Lillicrap, J.~Ba, and M.~Norouzi, ``Dream to control: Learning behaviors by latent imagination,'' \emph{arXiv preprint arXiv:1912.01603}, 2019.

\bibitem{hafner2020mastering}
D.~Hafner, T.~Lillicrap, M.~Norouzi, and J.~Ba, ``Mastering atari with discrete world models,'' \emph{arXiv preprint arXiv:2010.02193}, 2020.

\bibitem{hafner2023mastering}
D.~Hafner, J.~Pasukonis, J.~Ba, and T.~Lillicrap, ``Mastering diverse domains through world models,'' \emph{arXiv preprint arXiv:2301.04104}, 2023.

\bibitem{hansen2022temporal}
N.~Hansen, X.~Wang, and H.~Su, ``Temporal difference learning for model predictive control,'' \emph{arXiv preprint arXiv:2203.04955}, 2022.

\bibitem{hansen2023td}
N.~Hansen, H.~Su, and X.~Wang, ``Td-mpc2: Scalable, robust world models for continuous control,'' \emph{arXiv preprint arXiv:2310.16828}, 2023.

\bibitem{wu2023daydreamer}
P.~Wu, A.~Escontrela, D.~Hafner, P.~Abbeel, and K.~Goldberg, ``Daydreamer: World models for physical robot learning,'' in \emph{Conference on Robot Learning (CoRL)}, 2023.

\bibitem{shirwatkar2024pip}
A.~Shirwatkar, N.~Saxena, K.~Chandra, and S.~Kolathaya, ``Pip-loco: A proprioceptive infinite horizon planning framework for quadrupedal robot locomotion,'' \emph{arXiv preprint arXiv:2409.09441}, 2024.

\bibitem{nahrendra2024obstacle}
I.~Nahrendra, B.~Yu, M.~Oh, D.~Lee, S.~Lee, H.~Lee, H.~Lim, and H.~Myung, ``Obstacle-aware quadrupedal locomotion with resilient multi-modal reinforcement learning,'' \emph{arXiv preprint arXiv:2409.19709}, 2024.

\bibitem{silver2018residual}
T.~Silver, K.~Allen, J.~Tenenbaum, and L.~Kaelbling, ``Residual policy learning,'' \emph{arXiv preprint arXiv:1812.06298}, 2018.

\bibitem{yuan2024policy}
X.~Yuan, T.~Mu, S.~Tao, Y.~Fang, M.~Zhang, and H.~Su, ``Policy decorator: Model-agnostic online refinement for large policy model,'' \emph{arXiv preprint arXiv:2412.13630}, 2024.

\bibitem{rudin2022learning}
N.~Rudin, D.~Hoeller, P.~Reist, and M.~Hutter, ``Learning to walk in minutes using massively parallel deep reinforcement learning,'' in \emph{Conference on Robot Learning (CoRL)}, 2022.

\bibitem{wu2023learning}
J.~Wu, G.~Xin, C.~Qi, and Y.~Xue, ``Learning robust and agile legged locomotion using adversarial motion priors,'' \emph{IEEE Robotics and Automation Letters}, 2022.

\end{thebibliography}
